\documentclass{article}

    \PassOptionsToPackage{numbers, compress}{natbib}
\usepackage[preprint]{neurips_2026}

\usepackage[utf8]{inputenc} 
\usepackage[T1]{fontenc}    
\usepackage{hyperref}       
\usepackage{url}            
\usepackage{booktabs}       
\usepackage{amsfonts}       
\usepackage{nicefrac}       
\usepackage{microtype}      
\usepackage{xcolor}         
\usepackage{caption}
\usepackage{graphicx}
\usepackage[table]{xcolor}   
\usepackage{booktabs}   
\usepackage{amsmath} 
\usepackage{subcaption}
\usepackage{wrapfig}
\usepackage{makecell}

\title{WorldKV: Efficient World Memory with \\ World Retrieval and Compression}

\author{%
  Jung Yi$^{1}$\quad  Minjae Kim$^{1}$\quad  Paul Hyunbin Cho$^{1}$\quad  Wooseok Jang$^{1}$ \\
          \textbf{Sangdoo Yun}$^{2}$\quad \textbf{Seungryong Kim$^{1}$}\\[5pt]
  $^{1}$KAIST AI \quad  $^{2}$Naver AI Lab \\[5pt]
  {\tt\small\ \href{https://cvlab-kaist.github.io/WorldKV/}{https://cvlab-kaist.github.io/WorldKV/}} \\
}

\begin{document}

\maketitle

\vspace{-1.2em}
\begin{abstract}

Autoregressive video diffusion models have enabled real-time, action-conditioned world generation. However, sustaining a persistent world, where revisiting a previously seen viewpoint yields consistent content, remains an open problem. Full KV-cache attention preserves this consistency but breaks real-time constraints: memory footprint and attention cost grow linearly with rollout length.
Sliding-window inference restores throughput but discards long-term consistency. We propose WorldKV, a training-free framework with two components: World Retrieval and World Compression. World Retrieval stores evicted KV-cache chunks in GPU/CPU memory and selectively retrieves scene-relevant chunks via camera/action correspondence, inserting them back into the native attention window without re-encoding. World Compression prunes redundant tokens within each chunk via key-key similarity to an anchor frame, halving per-chunk storage to fit $2\times$ more history under a fixed budget. On Matrix-Game-2.0 and LingBot-World-Fast, WorldKV matches or exceeds full-KV memory fidelity at roughly 2$\times$ the throughput, and is competitive with memory-trained baselines without any fine-tuning.

\end{abstract}


\section{Introduction}
\label{sec:intro}

Autoregressive video diffusion models with causal attention and KV-caching have recently emerged as a promising architecture for real-time interactive world generation~\cite{team2026advancing, hong2025relic, sun2025worldplay, he2025matrix, seo2026grounding, ye2025yan, inspatioteam2026inspatioworldrealtime4dworld, savva2026solaris, gao2026dreamdojo}. These models generate action- or camera-conditioned visual streams at real-time frame rates, enabling applications in gaming~\cite{sun2025worldplay, hong2025relic}, embodied AI agents~\cite{inspatioteam2026inspatioworldrealtime4dworld}, and robotic simulation~\cite{gao2026dreamdojo, ye2026world}. Beyond producing plausible frames, the emerging goal is to sustain a persistent, explorable world — one in which a user can navigate freely, leave a room, and return to find it unchanged.

\begin{figure}
    \centering
    \includegraphics[width=1.0\linewidth]{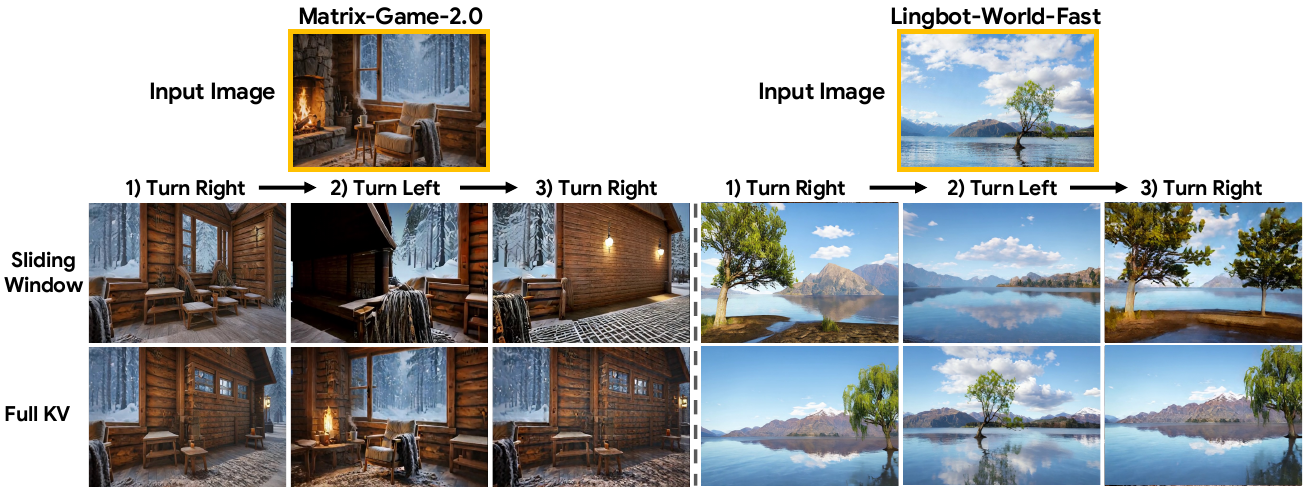}
    \caption{\textbf{Emergent memory from attending Full KV cache}: We empirically observed that, even though Matrix-Game-2.0~\cite{he2025matrix} was trained only on short clips, the model can use the KV cache as long-term visual context/memory.}
    \label{fig:emergent}
\end{figure}

Achieving this kind of persistence is tightly linked to \textit{spatial and temporal memory}: the ability to retain and recall scene content across time and revisits. Yet despite rapid progress in world model architectures, consistent memory remains an open challenge. A consistent world model should reconstruct the same structures and appearances when revisiting previously explored areas. However, models operating under sliding-window inference tend to hallucinate new content or drift~\cite{he2025matrix, savva2026solaris}, as the KV-caches from the original scene have long been evicted from the context.

A recent observation is that the KV cache in these models is not merely a computational buffer---it already functions as an emergent form of \textit{world memory}.  
LingBot-World~\cite{team2026advancing} demonstrated that, even without explicit memory training, attending to the full history of KV-caches enables the model to maintain spatial and temporal consistency across revisits. 
However, LingBot-World~\cite{team2026advancing} was trained on minute-level videos, so its long-term memory may reflect learned behavior rather than a property of the KV cache itself.
In this paper, we show that the phenomenon is more fundamental: 
it appears even in models not trained on long sequences. 
On Matrix-Game-2.0~\cite{he2025matrix}, which was trained on short sequences with a 6-frame sliding window, the model can nonetheless leverage past KV caches as long-term visual memory at inference time (Fig.~\ref{fig:emergent}). 
When we remove the sliding-window restriction and let the model attend to its entire KV-cache history, the model successfully reproduces previously seen viewpoints, while the same model under sliding-window inference fails.
The memory is already there; the question is how to access it without the full cost of attending to the entire KV cache.

        

\begin{figure}[t]
    \centering
    \captionsetup[subfigure]{skip=2pt}  
    \begin{subfigure}[b]{0.467\linewidth}
        \includegraphics[width=\linewidth]{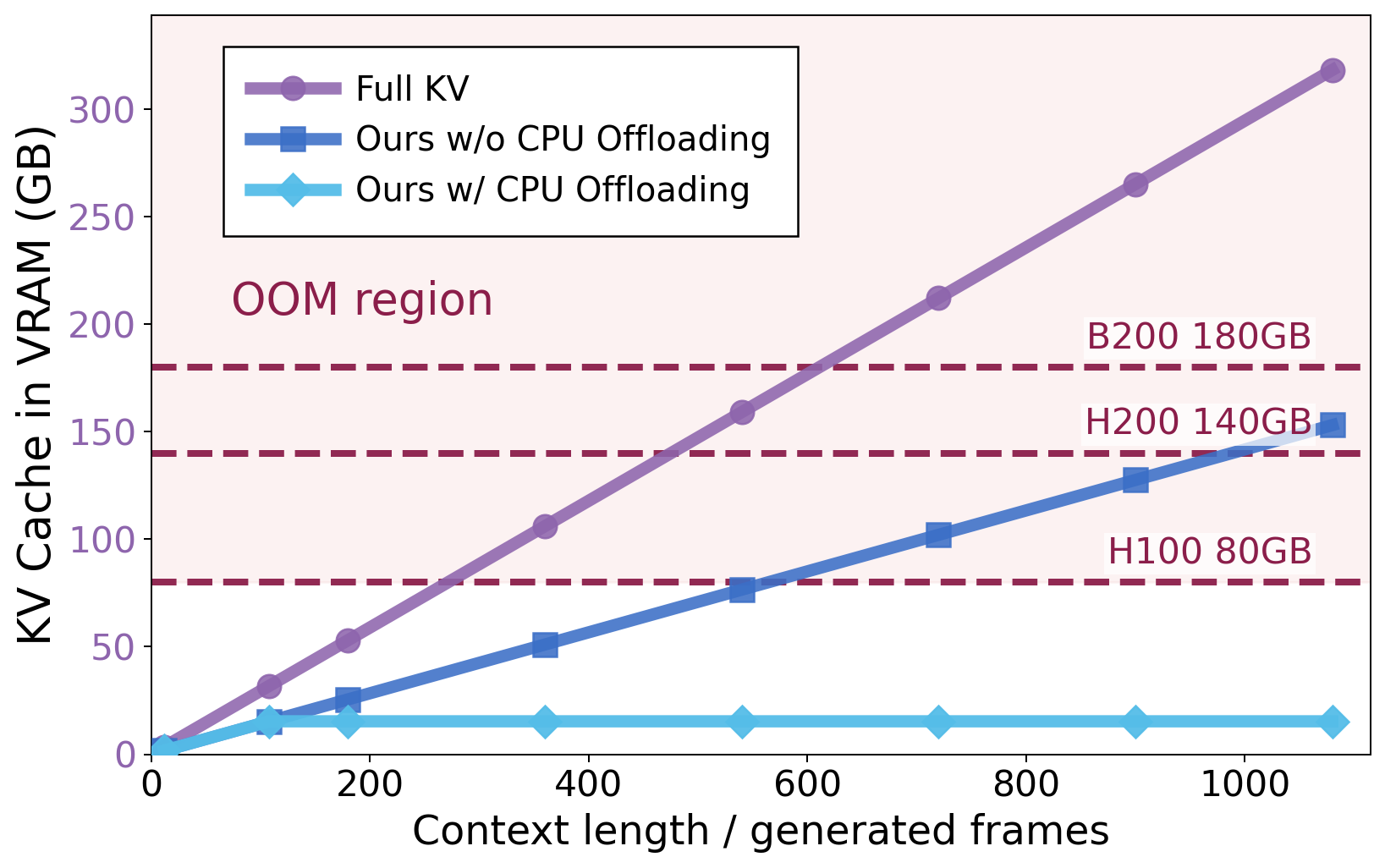}
        \caption{KV cache size (LingBot-World-Fast 14B)}
    \end{subfigure}
    \hfill
    \begin{subfigure}[b]{0.497\linewidth}
        \includegraphics[width=\linewidth]{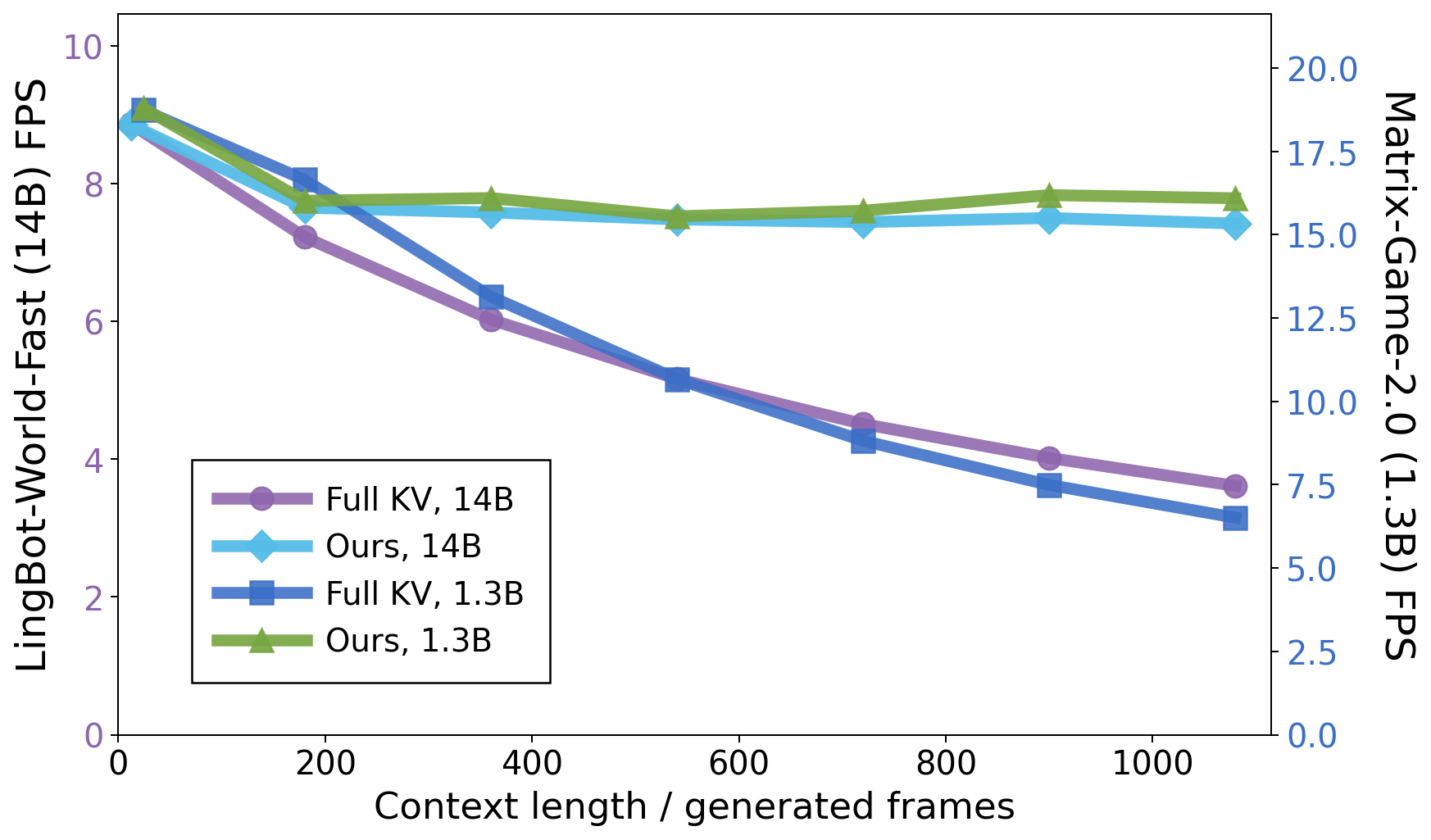}
        \caption{FPS over rollout length}
    \end{subfigure}
    \vspace{-5pt}
    \caption{\textbf{Cost of full-history KV-cache attention.} (a) Full KV grows rapidly into the OOM region, while WorldKV grows gradually via compression and stays nearly flat with CPU offloading. (b) Full KV throughput degrades continuously on both backbones (14B on 4×B200, 1.3B on 4×H200), while WorldKV maintains stable throughput.}
    \label{fig:cost}
\end{figure}

Indeed, the cost of leveraging this emergent memory through full-history KV cache attention is substantial in practice: each frame produces 880~\cite{he2025matrix} to 1{,}560~\cite{team2026advancing} tokens, accumulating hundreds of thousands of tokens over a one-minute rollout. The corresponding KV cache rapidly exceeds GPU VRAM capacity 
(Fig.~\ref{fig:cost} (a)). 
Even before out-of-memory failures, the rapidly growing attention cost degrades inference speed: on LingBot-World-Fast, FPS drops from 8.87 to 3.61 over a one-minute rollout 
(Fig.~\ref{fig:cost} (b)), breaking real-time constraints. Sliding-window inference is therefore a structural necessity for real-time generation, yet it comes with an inherent trade-off: the eviction that bounds attention cost is also what discards long-term memory.

Recent works address this through memory-augmented architectures: 
external memory banks retrieved via cross-attention~\cite{xiaoworldmem, huang2025memory}, 
spatial compression of the entire history~\cite{hong2025relic}, 
or explicit 3D scene representations~\cite{ren2025gen3c, li2025vmem} that condition the video model on rendered views of the reconstructed geometry.
While effective, these approaches require training dedicated memory modules or fine-tuning the backbone, and the 3D-representation methods additionally incur reconstruction latency at inference time.

We take a different perspective. Rather than building external memory on top of the model, we observe that the model's own KV cache is already available for world memory. 
We introduce \textbf{WorldKV}, a training-free framework that enables efficient long-term memory in autoregressive video world models through two complementary components: \textbf{World Retrieval} and \textbf{World Compression}.


\textbf{World Retrieval} 
preserves KV-cache chunks by storing them in GPU/CPU memory and selectively retrieving scene-relevant caches back into the active attention window when the model revisits a scene. Retrieved KV caches are inserted back into the context natively, with no re-encoding or architectural changes required. The retrieval mechanism is modular, supporting camera/action-based and attention-based strategies as interchangeable components (Appendix~\ref{sec:algorithm}).

\textbf{World Compression} reduces redundancy in adjacent frames, which 
produce near-duplicate KV caches. By pruning redundant tokens based on Key-Key similarity, each chunk is roughly
halved in size, allowing twice as much history under the same memory budget.
This preserves memory fidelity comparable to full KV-cache attention, and in
some cases even surpasses it.


Our contributions are as follows:
\begin{itemize}

\item We introduce \textbf{World Retrieval}, a retrieval-algorithm-agnostic framework that stores and selectively retrieves evicted KV-cache chunks, supporting camera/action-based and attention-based strategies as interchangeable components.


\item We present \textbf{World Compression}, a key-similarity-based pruning mechanism that compresses each chunk to approximately half its original size, enabling 2$\times$  more history under the same memory budget while preserving or improving revisit fidelity.

\item We quantitatively demonstrate, on two autoregressive video world models of different scales (Matrix-Game-2.0~\cite{he2025matrix}, LingBot-World-Fast~\cite{team2026advancing}), that training-free KV-cache management matches or exceeds both full KV-cache attention and memory-trained baselines on revisit fidelity while maintaining real-time inference.

\end{itemize}





\section{Related Work}
\label{sec:related}

\paragraph{Autoregressive Video Diffusion.}

Recent work~\cite{chen2024diffusion, huang2025self, teng2025magi, yin2025slow, zhu2026causal, liu2025rolling, chen2025skyreels} integrates diffusion modeling 
with autoregressive (AR) prediction for long-horizon and streaming video 
generation. 
CausVid~\cite{yin2025slow} distills a bidirectional diffusion transformer into a causal AR generator.
Self Forcing~\cite{huang2025self} mitigates mismatch between training and inference by training on self-generated rollouts with KV caching. Rolling Forcing~\cite{liu2025rolling} jointly denoises multiple frames at 
progressively increasing noise levels. LongLive~\cite{yang2025longlive} introduces KV re-caching for smooth prompt transitions. 
Building on this line of work, real-time interactive video world models leveraging KV caching have emerged as a natural extension, exploiting cached past states for low-latency generation under streaming user input.


\paragraph{Interactive World Model.}
Building on autoregressive video diffusion, interactive world models predict action-conditioned future frames. Matrix-Game-2.0~\cite{he2025matrix} 
injects keyboard and mouse signals, while 
Hunyuan-GameCraft~\cite{li2025hunyuangamecrafthighdynamicinteractivegame} 
unifies them into a  camera action space. Yume-1.5~\cite{mao2025yume}
further extends interactive exploration with text-controlled event generation, and LingBot-World~\cite{team2026advancing} scales interactive world generation toward diverse domains and long-horizon rollouts. A growing line of work has explored memory mechanisms for long-term consistency in world models.  
WorldPlay~\cite{sun2025worldplay} rebuilds context from geometrically important past frames via KV cache recomputation, with memory-aware distillation. RELIC~\cite{hong2025relic} introduces a learnable action-aware compression mechanism that stores historical latent memory in the KV cache.
In contrast, our framework operates training-free, exploiting sparse relevance and token redundancy in the existing KV cache.

\paragraph{KV Cache Management.} 

In autoregressive generation, the KV cache grows linearly with sequence length, creating a bottleneck for long-context inference. In LLMs, fixed-budget cache management has been studied through positional heuristics~\cite{xiao2023streamingllm}, accumulated attention scores~\cite{zhang2023h2o}, observation-window importance estimates~\cite{li2024snapkv, ghadia2025dialogue}, and query-aware page retrieval~\cite{tang2024quest}. 
While these methods reduce the cost of language-model decoding, they are not
designed for dense spatiotemporal generation. Recent work has begun to explore
training-free KV-cache management for long-horizon autoregressive video
diffusion~\cite{yi2025deep}. We extend this direction to interactive world
models, where long-horizon consistency further requires retrieving
scene-relevant memory across revisited viewpoints while compressing redundant
visual KV caches.





\begin{figure}[htbp]
    \centering
    \includegraphics[width=1.0\linewidth]{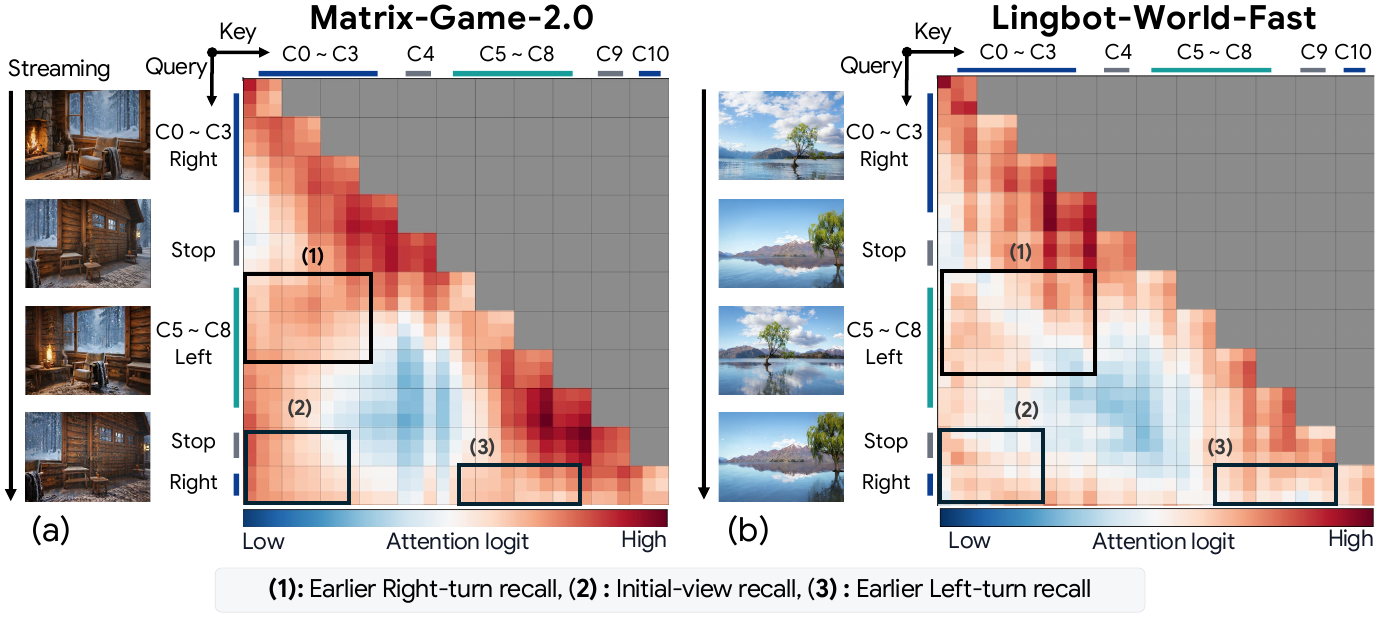}
    \caption{Attention maps for the action sequence ``Right (chunks C0--C3) $\rightarrow$
     Stop (C4) $\rightarrow$
     Left (C5--C8) $\rightarrow$
     Stop (C9) $\rightarrow$
     Right (C10)''. The model assigns high attention to KV-caches whose viewpoint overlaps with the current action. This pattern motivates using camera/action as the retrieval criterion for selecting relevant past KV caches.}
    \label{fig:attention_map}
\end{figure}

\section{Preliminaries}
\label{sec:preliminaries}
\paragraph{Interactive World Models.}
An interactive world model aims to predict future visual observations from actions. 
Given the current visual state $s_t \in \mathcal{S}$ and an action $a_t \in \mathcal{A}$, the model defines a conditional distribution over the next state $s_{t+1}$:
\begin{equation}
    s_{t+1} \sim p(s_{t+1} \mid s_t, a_t),
\end{equation}
where $p: \mathcal{S} \times \mathcal{A} \to \Delta(\mathcal{S})$ is the transition distribution.
In this work, ``world model'' refers specifically to this action-conditioned visual generation setting. Recent world models~\cite{he2025matrix, team2026advancing, savva2026solaris, inspatioteam2026inspatioworldrealtime4dworld} implement this transition using autoregressive video diffusion built on causal DiT architectures, conditioned on discrete keyboard actions or continuous camera trajectories.

\paragraph{Autoregressive Video Diffusion with KV Cache.}
Autoregressive video diffusion models~\cite{huang2025self, teng2025magi, chen2025skyreels} synthesize long
videos by sequentially generating frames or chunks (e.g., 3 frames). For a video of
$N$ frames $x^{1:N} = (x^1, x^2, \dots, x^N)$, the generation is
factorized as:
\begin{equation}
    p(x^{1:N}) = \prod_{i=1}^{N} p(x^i \mid x^{<i}),
\end{equation}
where each conditional is modeled by a diffusion process. In
practice, recent causal diffusion transformers~\cite{huang2025self,yang2025longlive, liu2025rolling} implement this conditioning
through a KV cache,
which stores key-value projections of previously generated
frames or chunks.
At step $t$, the transformer
$\mathcal{G}_\theta$ denoises a noisy latent $x_t^{(\sigma)}$
conditioned on prior cached entries:
\begin{equation}
    \hat{x}_t = \mathcal{G}_\theta(x_t^{(\sigma)},\; \sigma,\;
    \mathbf{K}_{<t},\; \mathbf{V}_{<t}).
\end{equation}
The new key-value pairs are appended to the cache for subsequent
steps. 
 
\section{Method}
\label{sec:method}
\subsection{Overview}
\label{sec:overview}
Our framework, \textbf{WorldKV}, operates on top of sliding-window inference and introduces two complementary components addressing the two bottlenecks of full-KV inference: attention computation and storage (Fig.~\ref{fig:cost} (a), (b)). \textbf{World Retrieval} (Sec.~\ref{sec:retrieval}) stores evicted KV-cache chunks in GPU/CPU memory and retrieves only viewpoint-relevant caches at revisit time, bounding the active attention window to preserve real-time inference speed. \textbf{World Compression} (Sec.~\ref{sec:compression}) prunes redundant tokens within each chunk via key-key similarity, compressing each 3-frame chunk to approximately half its size and fitting roughly 2$\times$ chunks under a fixed memory budget without out-of-memory failures.

\subsection{World Retrieval}
\label{sec:retrieval}

\begin{figure}[htbp]
    \centering
    \includegraphics[width=1.0\linewidth]{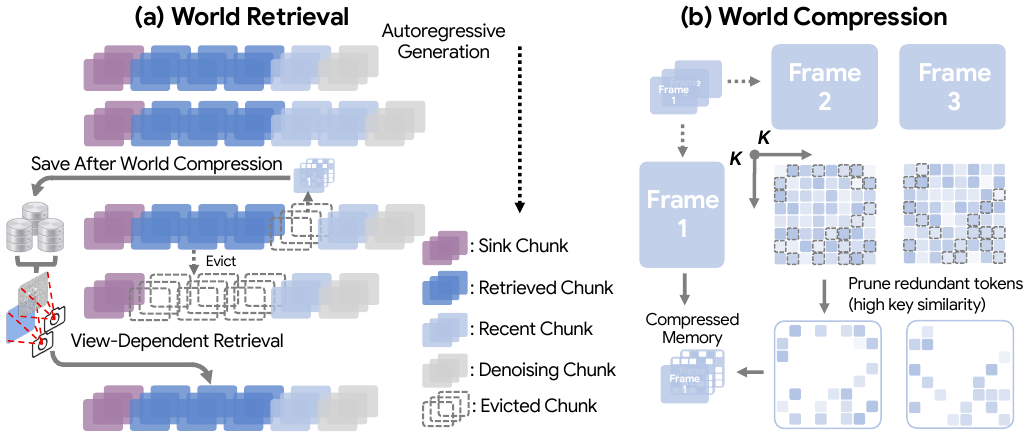}
    \caption{\textbf{Overview of WorldKV.} (a)~World Retrieval stores KV-cache chunks after compression and retrieves view-relevant chunks back into the attention window at revisit time. (b)~World Compression designates first frame of each chunk as the anchor, computes key similarity against the remaining frames, and prunes tokens redundant with the anchor.}
    \label{fig:main_method}
\end{figure}

\paragraph{Attention Sparsity under Camera/Action Revisits.}
We first analyze how autoregressive world models distribute attention over historical KV caches under camera/action input. We generate a sequence of 11 chunks following the trajectory ``Right (chunks C0--C3) $\rightarrow$
 Stop (C4) $\rightarrow$
 Left (C5--C8) $\rightarrow$
 Stop (C9) $\rightarrow$
 Right (C10)'' and visualize the chunk-level attention maps for Matrix-Game-2.0~\cite{he2025matrix} and LingBot-World-Fast~\cite{team2026advancing} in Fig.~\ref{fig:attention_map}.
The maps reveal a clear view-correspondence pattern across both models. As the camera turns Left at C5--C8 and sweeps back toward the initial scene direction, attention rises on C0--C2, whose cached views overlap with the current viewpoint, as indicated by (1).
At C9, where the camera stays near the initial viewpoint, attention concentrates on C0, the input image corresponding to that view, as marked by (2).
When the camera turns right again at C10, attention shifts toward C5--C8, the chunks generated during the previous left-turn trajectory, as highlighted by (3).
These patterns show that the model does not simply attend to the most recent caches; instead, it reuses past KV chunks whose viewpoints correspond to the current frame.
This observation suggests that attending to a compact set of viewpoint-relevant KV chunks can preserve
much of the important context provided by full-KV attention, motivating
\textbf{World Retrieval}.

\paragraph{World Retrieval Mechanism.}
Motivated by the view-correspondence pattern observed above, World Retrieval operates as follows. Under sliding-window inference, KV-cache chunks evicted from the active attention window are stored in GPU/CPU memory rather than discarded, each indexed by the camera/action state $\mathbf{a}_i$ at the time of its generation (absolute pose for camera models, cumulative discrete actions for keyboard action models).

As illustrated in Fig.~\ref{fig:main_method}(a), the sliding window is partitioned into four regions: 1) \textit{sink} KV caches from the initial frames that serve as a visual anchor, 2) \textit{retrieved} KV caches selected from stored history, 3) \textit{recent} KV caches from the immediately preceding frames, and 4) \textit{denoising} chunk currently being generated. World Retrieval operates on the retrieved region: at generation time, given the current camera/action state $\mathbf{a}_\text{cur}$, the top-$k$ most relevant chunks are selected from the stored history to fill this region:
\begin{equation}
    \mathcal{R} = \operatorname{Top\text{-}k} \left( \operatorname{sim}(\mathbf{a}_\text{cur}, \mathbf{a}_i) \mid i = 1, \dots, M \right),
\end{equation}
where $M$ is the number of stored chunks, $k$ is the retrieval budget, and $\operatorname{sim}(\cdot, \cdot)$ is a relevance function.

The framework is \textit{retrieval-algorithm agnostic}: $\operatorname{sim}(\cdot, \cdot)$ can be instantiated as camera/action-based similarity, query-based importance score, or other relevance methods. 
In this work, we evaluate camera/action-based and query-based retrieval in Appendix~\ref{sec:algorithm}; both substantially outperform sliding-window inference, demonstrating that the framework generalizes across retrieval signals.

\subsection{World Compression}
\label{sec:compression}

\paragraph{Motivation.}

World Retrieval requires storing all evicted KV caches in GPU/CPU memory for potential future retrieval. However, this storage cost is substantial: on LingBot-World-Fast, a single chunk of 3 latent frames occupies approximately 3.4GB across all transformer layers, accumulating to over 200GB for a one-minute rollout — exceeding even the VRAM capacity of a B200 GPU (Fig.~\ref{fig:cost} (a)). We observe that temporally adjacent frames share substantial visual content (Appendix~\ref{sec:keykey}): camera viewpoint, scene layout, and object appearance change minimally over consecutive frames, producing near-duplicate KV caches that encode largely overlapping information. World Compression exploits this redundancy to reduce per-chunk storage while preserving the most distinctive KV caches. Beyond storage savings, this enables broader retrieval coverage within a fixed attention budget; as we show in Sec.~\ref{sec:main_ablation} and Appendix~\ref{sec:increasing}, broader coverage improves revisit fidelity.
 
\paragraph{Key-Key Similarity as a Redundancy Measure.}
World Compression requires a criterion for identifying redundant 
tokens within a short temporal chunk. We use Key-Key cosine 
similarity as a redundancy signal: we compare non-anchor frame 
keys against the anchor-frame keys, and find that keys from 
spatiotemporally overlapping regions exhibit high cosine similarity 
while keys from newly revealed or dynamic regions diverge 
(Appendix~\ref{sec:keykey}). This finding is also consistent with 
prior evidence that keys in video diffusion transformers encode 
spatiotemporal correspondence~\cite{yuan2025denoise}. We therefore 
prune high-similarity non-anchor tokens as redundant with the 
anchor, while retaining low-similarity tokens that carry 
distinctive content.

\paragraph{World Compression Mechanism.}
Given a chunk consisting of $F$ consecutive frames, World Compression designates the first frame as the \textit{anchor} and compresses the remaining $F-1$ frames against it.
Concretely, let $\mathbf{K}^{(a)} \in \mathbb{R}^{T \times d}$ denote the $T$ key vectors from the anchor frame at a given layer. 

For each non-anchor frame $f$, we measure the redundancy of each key $\mathbf{k}_j^{(f)}$ as its average cosine similarity to all anchor-frame keys:
\begin{equation}
    s_j^{(f)} =
    \frac{1}{T}\sum_{i=1}^{T}
    \frac{{\mathbf{k}_j^{(f)}}^\top \mathbf{k}_i^{(a)}}
    {\|\mathbf{k}_j^{(f)}\| \cdot \|\mathbf{k}_i^{(a)}\|}.
\end{equation}

We pool these scores across all non-anchor frames and retain the bottom $P\%$ among the pooled non-anchor tokens by similarity, since low similarity indicates content not captured by the anchor, such as newly revealed regions under camera motion.
The compressed chunk consists of all anchor-frame tokens plus the retained tokens. With $F=3$ and $P=25\%$ retention across the $2T$ non-anchor tokens, each chunk shrinks from $3T$ to approximately $1.5T$ tokens, achieving $2\times$ storage efficiency.

Compression is applied once per chunk at storage time and operates independently per layer: each layer retains its own set of distinctive tokens, since token importance varies across layers. At retrieval time, each layer attends to its own retained tokens within the inserted chunk. Beyond storage efficiency, compression improves revisit fidelity by reducing redundancy in the attention window; we analyze this in Sec.~\ref{sec:main_ablation}.

\section{Experiments}
\label{sec:experiments} 

\subsection{Experimental settings}
\label{sec:settings}

\paragraph{Benchmark.}

To evaluate the memory performance of world models, we construct a benchmark of 60 scene-trajectory pairs spanning diverse visual domains (e.g., indoor, outdoor, urban, natural). Initial frames are sourced from real-world videos, game recordings, and AI-generated images. For each scene, we manually design a long-horizon trajectory containing diverse camera/action sequences — repetitive revisits, forward-backward traversals, and their combinations — with at least one loop-closure event where the camera returns to a previously observed viewpoint, enabling direct evaluation of revisit consistency.
 
\paragraph{Base Models.}
We evaluate on two autoregressive video world models at different scales: (1)~LingBot-World-Fast~\cite{team2026advancing} is a 14B-parameter model distilled from a long-video teacher capable of generating one-minute sequences; it natively operates with full KV-cache attention. (2)~Matrix-Game-2.0~\cite{he2025matrix} is a 1.3B-parameter model that was not trained on long-context video; it natively operates with a sliding window of 6 latent frames.


\paragraph{Baselines.}

For each base model, we compare against its native inference mode (full KV-cache attention for LingBot-World-Fast~\cite{team2026advancing}, sliding-window inference for Matrix-Game-2.0~\cite{he2025matrix}). We additionally compare against WorldPlay~\cite{sun2025worldplay} and Yume-1.5~\cite{mao2025yume}, which were trained with memory modules.

\paragraph{Implementation Details.}
We use a sliding window of 18 latent frames partitioned into sink (3 frames), retrieval (9 frames), recent (3 frames), and denoising (3 frames). World Compression retains the anchor frame in full and keeps 25\% of tokens in non-anchor frames, compressing each 3-frame chunk to 1.5 frames. 

For retrieval, we adopt a unified camera/action-based strategy across both models. Each evicted KV chunk is stored alongside its camera translation and rotation. At retrieval time, we compute a combined distance from the squared L2 distance in translation and geodesic distance in rotation, each normalized across the retrieval candidate set, and sum the two to form the final retrieval distance; chunks with the smallest distance are retrieved. For LingBot-World-Fast~\cite{team2026advancing}, camera poses are directly available.  For Matrix-Game-2.0~\cite{he2025matrix}, which accepts discrete keyboard and mouse inputs, we accumulate WASD and yaw/pitch commands into pseudo-translation and pseudo-rotation vectors. While these are not calibrated to scene geometry, they capture the relative camera motion induced by the action sequence, making them effective for retrieval.
 
\paragraph{Metrics.}
We measure PSNR, SSIM~\cite{Wang2004ImageQA}, and LPIPS~\cite{Zhang2018TheUE} between each revisit frame and the corresponding first-visit frame generated at the same viewpoint. For FID~\cite{Heusel2017GANsTB}, we compute the distributional distance between the set of revisit frames and the set of first-visit reference frames. Higher PSNR/SSIM and lower LPIPS/FID indicate better memory fidelity. We also report throughput (FPS) measured at the last chunk of each rollout.

\subsection{Quantitative results}

\begin{table}[t]
\centering
\small
\caption{
\textbf{Quantitative comparison on world memory evaluation.} 
We apply WorldKV to  video world models, 
measuring revisit consistency between 
revisited and first-visit frames.
Throughputs are measured on 4$\times$H200 GPUs by default; values 
in () are measured on 4$\times$B200 GPUs.
}

\label{tab:main}
\begin{tabular}{l|c|cccc}
\toprule
\textbf{Method} & \makecell{\textbf{Throughput} $\uparrow$ \\ (FPS)}  & \textbf{LPIPS} $\downarrow$ & \textbf{PSNR} $\uparrow$ & \textbf{SSIM} $\uparrow$ & \textbf{FID} $\downarrow$ \\
\midrule
\rowcolor{gray!10}
\multicolumn{6}{l}{\textit{With Memory Training}} \\

WorldPlay (8B)  & 4.95   & 0.496   & 14.556  & 0.470   & 113.317 \\
Yume-1.5 (5B)   & 11.41   & 0.566   & 13.167   & 0.467   & 141.704 \\

\midrule
\rowcolor{gray!10}
\multicolumn{6}{l}{\textit{Without Memory Training}} \\

\multicolumn{6}{l}{\textbf{LingBot-World-Fast} (14B)} \\
~~- Sliding Window     & 5.05 (8.10)   & 0.581   & 12.184   & 0.375   & 144.036 \\
~~- Full KV (Original)     & 2.36 (3.83)    & \textbf{0.441}   & \textbf{15.901}   &  \textbf{0.472}   & \underline{85.705} \\
~~- \textbf{WorldKV(Ours)}      & 4.78 (7.71)   & \underline{0.455} & \underline{15.660} & \underline{0.463} & \textbf{75.644} \\
\multicolumn{6}{l}{\textbf{Matrix-Game-2.0} (1.3B)} \\
~~- Sliding Window (Original)           & 18.87   & 0.594   & 11.422   & 0.280   & 157.261 \\

~~- Full KV      & 7.82   & \underline{0.529}   & \underline{13.748}   & \underline{0.364}   & \underline{124.912} \\
~~- \textbf{WorldKV(Ours)}      & 16.25   & \textbf{0.462}   &  \textbf{14.101}   & \textbf{0.405} & \textbf{93.561} \\

\bottomrule
\end{tabular}
\vspace{4mm}
\end{table}
 
 
 

Table~\ref{tab:main} summarizes the main comparison. Sliding-window baselines perform poorly on both models, as evicted caches leave the model with no access to previously generated scene content at revisit time. In contrast, WorldKV maintains throughput close to sliding-window inference, while full KV-cache attention drops to less than half this throughput due to linearly growing context length.

On LingBot-World-Fast~\cite{team2026advancing}, which was distilled from a long-video teacher, full KV-cache attention already provides strong memory. WorldKV closely approaches Full KV performance across all metrics at roughly 2 $\times$ the throughput.

On Matrix-Game-2.0~\cite{he2025matrix}, WorldKV outperforms both sliding-window and full KV-cache attention across all metrics. Full KV underperforms here because Matrix-Game-2.0 was trained on short sequences: the accumulated KV cache contains degraded KV caches from out-of-distribution generation, and attending to all of them introduces compounding errors~\cite{yi2025deep}. WorldKV avoids this by retrieving only the KV caches relevant to the current scene.



Compared to memory-trained baselines, LingBot-World-Fast~\cite{team2026advancing} with WorldKV outperforms WorldPlay~\cite{sun2025worldplay} and Yume-1.5~\cite{mao2025yume} on LPIPS, PSNR, and FID with comparable SSIM, despite requiring no memory-specific training. Matrix-Game-2.0~\cite{he2025matrix} with WorldKV achieves competitive performance against memory-trained baselines.

\subsection{Qualitative results}
\label{sec:qual}

\begin{figure}[htbp]
    \centering
    \includegraphics[width=0.97\linewidth]{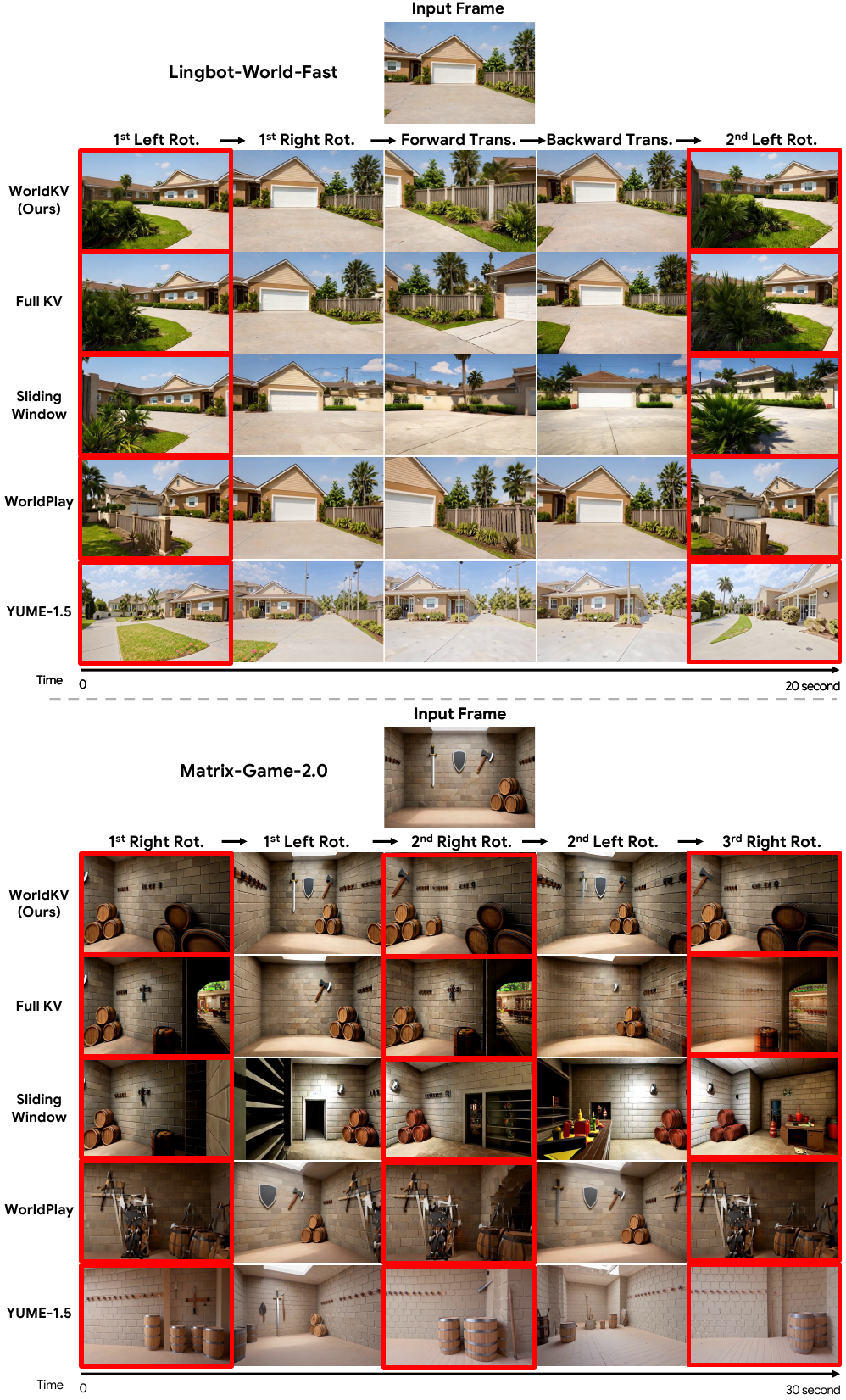}
    \caption{
    Frame-by-frame comparison across methods on two trajectories.
    }
    \label{fig:main_qual}
\end{figure}

Fig.~\ref{fig:main_qual} shows generations on two scenes under multi-revisit trajectories. 
On LingBot-World-Fast, WorldKV closely matches Full KV, recovering scene appearance with high fidelity, and Appendix~\ref{sec:worldvsfull} shows cases where WorldKV reconstructs revisited scenes more faithfully than Full KV. On Matrix-Game-2.0, where Full KV degrades over long horizons, WorldKV produces sharper and more consistent results than Full KV. Sliding-window drifts visibly on both backbones, as evicted caches prevent access to the original viewpoint. Compared to memory-trained baselines~\cite{sun2025worldplay, mao2025yume}, WorldKV clearly outperforms them on LingBot-World-Fast and remains comparable on Matrix-Game-2.0. 

\subsection{Ablation studies}
\label{sec:main_ablation}

We ablate World Compression from two perspectives. First, the upper part of 
Table~\ref{tab:compression_ablation} varies the \textit{intra-chunk} compression 
ratio, i.e., how many frame-equivalents are retained from each 3-frame chunk. 
The notation $3\to r$ means that a 3-frame chunk is compressed to $r$ 
frame-equivalents, ranging from anchor-only ($3\to1.0$) to no compression 
($3\to3.0$); the latter corresponds to applying only World Retrieval(WR). Second, the lower part varies the \textit{inter-chunk} coverage 
under a fixed 3-chunk-equivalent budget. For example, $6\to3$ compresses 6 
chunks into the same budget as 3 chunks, increasing history 
coverage without enlarging the attention window; the $3\to3$ row again 
corresponds to the WR-only baseline. Additional ablations on the 
number of retrieved chunks are provided in Appendix~\ref{sec:increasing}.

\paragraph{Intra-Chunk Compression Ratio Ablation.}
\label{sec:main_ablation_intra}
Table~\ref{tab:compression_ablation} (Top) shows the effect of varying the compression ratio within each chunk. On both models, retaining only the anchor frame ($3\to1.0$) yields lower performance, confirming that non-anchor tokens contain distinctive information not captured by the anchor alone. Moderate compression ($3\to1.5$ or $3\to2.0$) achieves strong performance, while retaining more tokens gives limited gains. This suggests that World Compression preserves informative entries at practical compression ratios.

\begin{table}[t]
\centering
\small
\caption{Ablation on World Compression. (Top) Intra-chunk compression (frames retained per 3-frame chunk). (Bottom) Inter-chunk compression (chunks compressed into a fixed 3-chunk budget).}
\label{tab:compression_ablation}
\renewcommand{\arraystretch}{0.9}
\begin{subtable}{\linewidth}
\centering
\label{tab:compression_intra}
\resizebox{\linewidth}{!}{%
\begin{tabular}{lcccccccccccc}
\toprule
& \multicolumn{4}{c}{\textbf{LingBot-World-Fast}}
& \multicolumn{4}{c}{\textbf{Matrix-Game-2.0}} \\
\cmidrule(lr){2-5}\cmidrule(lr){6-9}
\textbf{Intra-Chunk}
& \textbf{LPIPS}$\downarrow$ & \textbf{PSNR}$\uparrow$ & \textbf{SSIM}$\uparrow$ & \textbf{FID}$\downarrow$
& \textbf{LPIPS}$\downarrow$ & \textbf{PSNR}$\uparrow$ & \textbf{SSIM}$\uparrow$ & \textbf{FID}$\downarrow$ \\
\midrule
$3 \to 1.0$  & 0.494 & 14.807 & 0.435 & 97.155 & 0.474 & 13.951 & 0.385 & 94.424 \\
$3 \to 1.25$ & 0.463 & 15.599 & 0.467 & 90.124 & 0.463 & 14.158 & 0.397 & 95.376 \\
$3 \to 1.5$  & 0.455 & 15.660 & 0.463 & 75.644 & 0.462 & 14.101 & 0.405 & 93.561 \\
$3 \to 2.0$  & 0.456 & 15.654 & 0.461 & 76.230 & 0.453 & 14.258 & 0.417 & 96.000 \\
$3 \to 2.5$  & 0.456 & 15.685 & 0.466 & 77.482 & 0.450 & 14.216 & 0.417 & 95.478 \\
$3 \to 3.0$  & 0.459 & 15.594 & 0.467 & 75.005 & 0.443 & 14.630 & 0.433 & 93.536 \\
\bottomrule
\end{tabular}%
}
\end{subtable}

\vspace{0.2em}

\begin{subtable}{\linewidth}
\centering
\label{tab:compression_inter}
\resizebox{\linewidth}{!}{%
\begin{tabular}{lcccccccccccc}
\toprule
& \multicolumn{4}{c}{\textbf{LingBot-World-Fast}}
& \multicolumn{4}{c}{\textbf{Matrix-Game-2.0}} \\
\cmidrule(lr){2-5}\cmidrule(lr){6-9}
\textbf{Inter-Chunk}
& \textbf{LPIPS}$\downarrow$ & \textbf{PSNR}$\uparrow$ & \textbf{SSIM}$\uparrow$ & \textbf{FID}$\downarrow$
& \textbf{LPIPS}$\downarrow$ & \textbf{PSNR}$\uparrow$ & \textbf{SSIM}$\uparrow$ & \textbf{FID}$\downarrow$ \\
\midrule
$3 \to 3$  & 0.468 & 15.436 & 0.454 & 91.389 & 0.496 & 13.416 & 0.369 & 105.988 \\
$6 \to 3$  & \textbf{0.455} & \textbf{15.660} & \textbf{0.463} & \textbf{75.644} & \textbf{0.462} & \textbf{14.101} & \textbf{0.405} & \textbf{93.561} \\
$9 \to 3$  & 0.482 & 14.982 & 0.430 & 101.226 & 0.499 & 13.403 & 0.360 & 108.460 \\
\bottomrule
\end{tabular}%
}
\end{subtable}
\end{table}

\paragraph{Inter-Chunk Compression Ratio Ablation.}
Table~\ref{tab:compression_ablation} (Bottom) compares compression scopes under a fixed 3-chunk retrieval budget. On both models, $6$ chunks $\to$ $3$ chunks outperforms uncompressed retrieval ($3\to3$), showing that broader historical coverage is more useful than preserving fewer chunks at full resolution. This agrees with Appendix~\ref{sec:increasing}, where retrieving more chunks improves memory fidelity. In contrast, aggressive compression ($9\to3$) degrades performance by retaining only anchor frames and discarding distinctive non-anchor information.

\section{Conclusion}
\label{sec:conclusion}
We introduced WorldKV, a training-free framework for efficient world memory in autoregressive video world models through KV-cache retrieval and compression. WorldKV enables consistent scene revisits while maintaining real-time inference, achieving memory fidelity competitive with full KV-cache attention and memory-trained baselines across two world models of different scales, all without any fine-tuning or distillation and at  lower memory and attention cost.

\section{Limitations \& Future Work}
\label{sec:limitations}


WorldKV improves the efficiency and fidelity of long-horizon world-model
inference, while remaining an inference-time memory-management method. Since it
operates on the KV cache of a frozen backbone, its visual fidelity remains
bounded by the generation quality of the underlying pretrained world model. In
rollouts substantially longer than those seen during training, autoregressive
video generation may still accumulate visual artifacts caused by error accumulation. WorldKV focuses on efficient preservation and
retrieval of past visual memory, and future work may combine it with training
strategies for more stable multi-minute world generation.


As shown in Fig.~\ref{fig:cost} (a), CPU offloading offers a complementary direction for reducing VRAM cost: all KV caches are stored in CPU memory, and only the chunks needed for current attention are loaded onto the GPU. This bounds VRAM consumption regardless of rollout length.
However, the host-device transfer
latency at retrieval time currently prevents real-time generation. We leave
reducing this offloading latency to future work, which would enable real-time
multi-minute world generation under bounded VRAM.

\clearpage
{
\small

\bibliographystyle{plainnat}
\bibliography{neurips_2026}
}
\clearpage


\appendix

\section*{Appendix}
\section{Selective Retrieval Outperforms Full KV Cache Attention}
\label{sec:worldvsfull}
\begin{figure}[htbp]
    \centering    \includegraphics[width=1.0\linewidth]{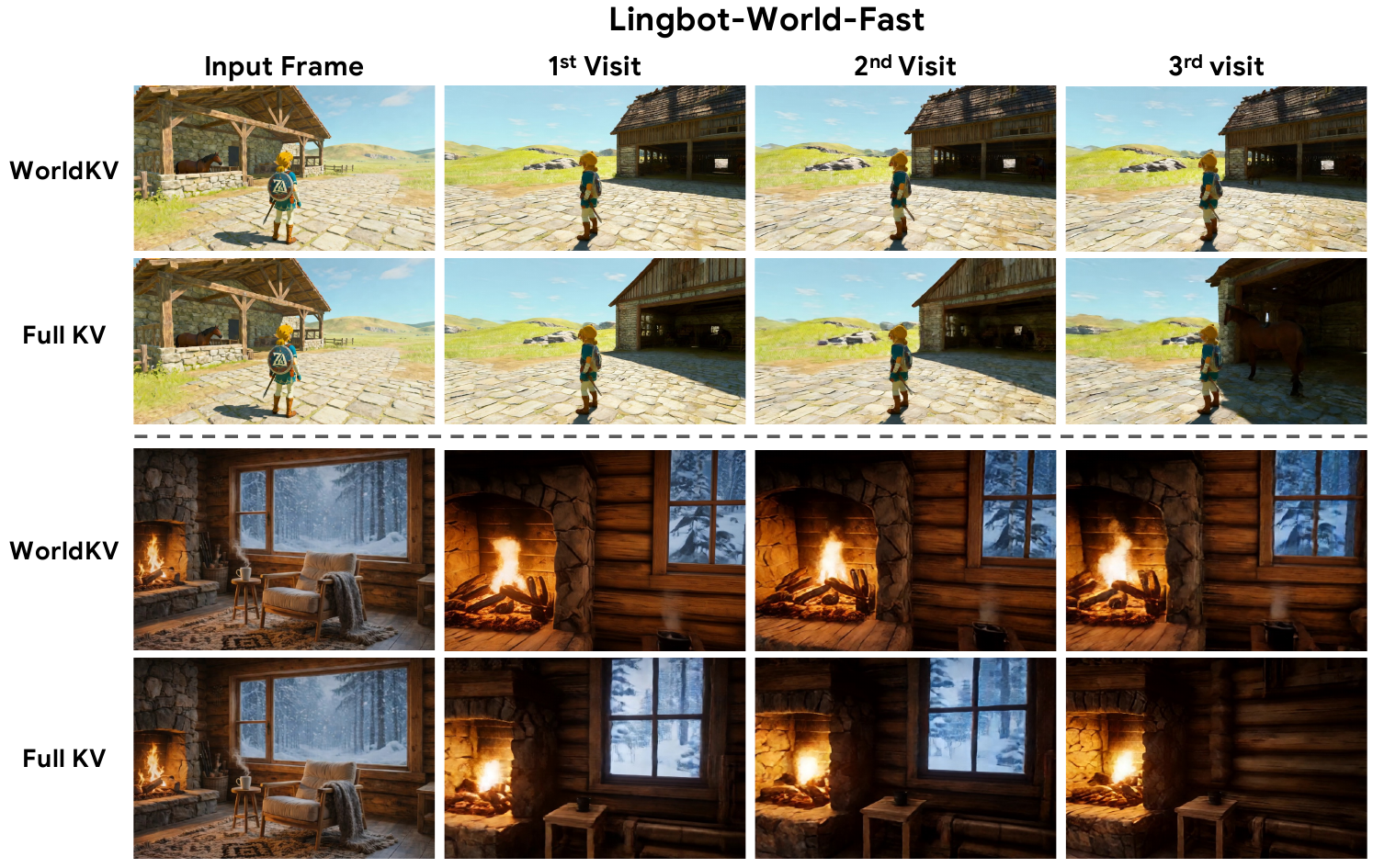}
    \caption{
    Each column shows the same viewpoint revisited at different times during a long-horizon rollout. Compared with full KV-cache attention, WorldKV preserves scene-specific details more faithfully across repeated visits by selectively retrieving viewpoint-relevant chunks and pruning redundant caches within each chunk.
    }
    \label{fig:worldvsfull}
\end{figure}

While full KV-cache attention generally provides strong memory fidelity on LingBot-World-Fast~\cite{team2026advancing}, Fig.~\ref{fig:worldvsfull} shows cases where WorldKV reconstructs revisited viewpoints more faithfully than attending to the full history. We hypothesize that this is related to attention dilution: as the KV cache grows, the model must attend over many entries, including redundant or viewpoint-irrelevant ones, which may weaken effective access to the entries that encode the revisited scene.
This interpretation is motivated by related findings in long-context language models. Lost in the Middle~\cite{liu-etal-2024-lost} shows that relevant information can be underutilized even when it is present in long contexts.
Similarly, KV-cache compression methods such as SnapKV~\cite{li2024snapkv} and R-KV~\cite{cai2025r} show that compact caches can match, and in some cases outperform, full-cache attention by reducing irrelevant or redundant context.
Although these results are obtained in language-model decoding, they suggest that full-history attention is not always optimal when much of the context is irrelevant or redundant.

In our setting, WorldKV follows this principle in dense spatiotemporal memory: it restricts the active attention window to retrieved viewpoint-relevant chunks and prunes near-duplicate entries within each chunk. To our knowledge, this is the first empirical observation of selective KV retrieval and compression outperforming full KV-cache attention in autoregressive video world models.
These cases suggest that selective retrieval and compression can provide a cleaner effective context than full-history attention in some long-horizon world-model rollouts.

\section{Key-Key Similarity Visualization}
\label{sec:keykey}
\begin{figure}[htbp]
    \centering    \includegraphics[width=0.75\linewidth]{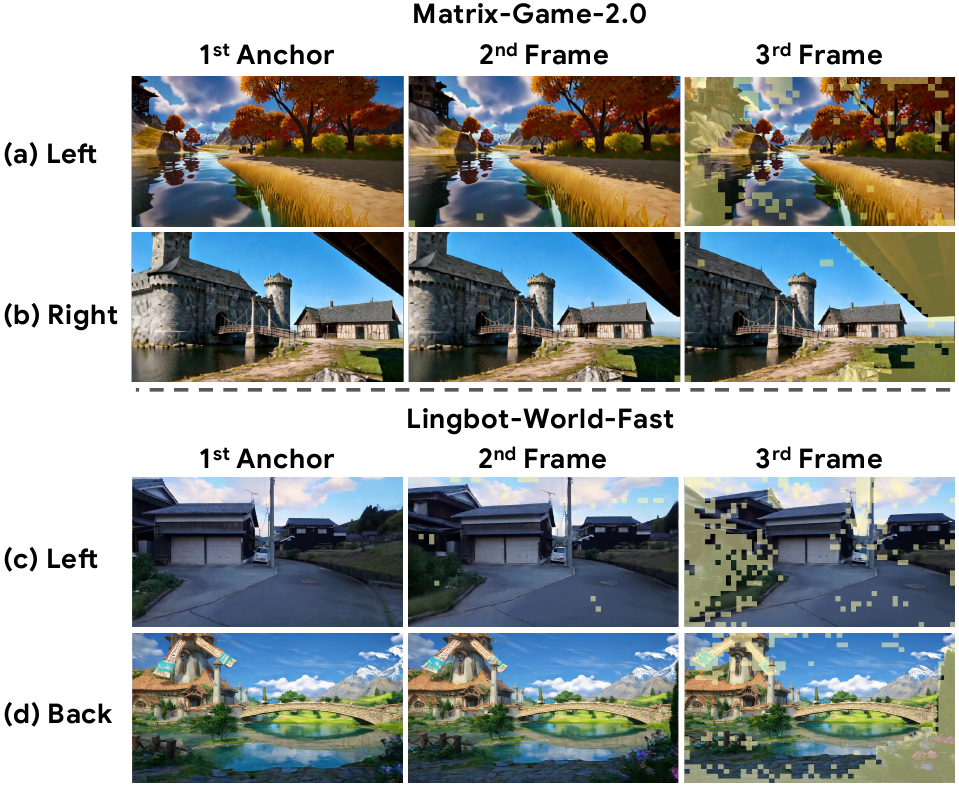}
    \caption{\textbf{Key-Key similarity visualization on Matrix-Game-2.0 and LingBot-World-Fast.} Yellow patches indicate tokens in the 2nd and 3rd frames with the lowest 12.5\% cosine similarity to the anchor-frame keys. These low-similarity tokens correspond to information not present in the anchor, including newly revealed regions under camera motion and dynamic object changes, and are retained by World Compression.
    }
    \label{fig:keykey}
\end{figure}

Figure~\ref{fig:keykey} visualizes the distinctive tokens selected by
Key-Key similarity. For each scene, we use the first frame as the anchor and
highlight the bottom 12.5\% least-similar tokens from the 2nd and 3rd frames,
based on cosine similarity to the anchor-frame keys. In camera-rotation cases
(Fig.~\ref{fig:keykey}(a), (b), (c)), low-similarity tokens concentrate on newly
revealed regions absent from the anchor frame, such as the left or right image
boundaries exposed by camera motion. This indicates that high key similarity
corresponds to redundant visual content, while low similarity identifies
newly visible information that should be preserved.

Beyond newly revealed spatial content, Key-Key similarity also captures dynamic
changes. In Fig.~\ref{fig:keykey}(d), where the camera moves backward,
low-similarity tokens appear not only on newly visible scene regions but also on
the rotating blades of the windmill. This suggests that key-similarity-based
pruning can preserve distinctive temporal information from moving objects,
rather than only static spatial differences. Overall, these observations support
Key-Key similarity as an effective cue for redundancy within a chunk.

\section{Retrieval Algorithm Ablations}
\label{sec:algorithm} 

As described in Sec.~\ref{sec:retrieval}, World Retrieval is
retrieval-algorithm agnostic. Table~\ref{tab:algorithm_ablation} compares two
retrieval strategies against the sliding-window baseline. The first is the
camera/action-based retrieval used in our main experiments
(Sec.~\ref{sec:settings}), which selects chunks by camera pose or accumulated discrete action
similarity. The second is query-based retrieval, which ranks stored chunks by
their attention scores with respect to the current denoising query, inspired by Deep Forcing~\cite{yi2025deep}. Both strategies substantially outperform
sliding-window inference on both models, indicating that useful scene memory can
be recovered from stored KV history across different retrieval signals.
Camera/action-based retrieval consistently performs best, suggesting that
viewpoint correspondence provides a particularly strong retrieval signal for
interactive world models. We therefore use camera/action-based retrieval as our
default and leave improved retrieval algorithms for future work.


\begin{table}[t]
\centering
\small
\caption{Retrieval Algorithm Ablation}
\label{tab:algorithm_ablation}
\begin{tabular}{lcccc}
\toprule
\textbf{Method} & \textbf{LPIPS} $\downarrow$ & \textbf{PSNR} $\uparrow$ & \textbf{SSIM} $\uparrow$ & \textbf{FID} $\downarrow$ \\
\midrule
\multicolumn{5}{l}{\textbf{LingBot-World-Fast}} \\
~~Sliding          & 0.581   & 12.184   & 0.375   & 144.036 \\
~~Camera / Action-based          & 0.455 & 15.660 & 0.463 & 75.644 \\
~~Query-based         & 0.490   & 15.065   & 0.445   & 83.201 \\
\multicolumn{5}{l}{\textbf{Matrix-Game-2.0}} \\
~~Sliding            &  0.594   & 11.422   & 0.280   & 157.261 \\
~~Camera / Action-based          & 0.462 & 14.101 & 0.405 & 93.561 \\
~~Query-based        & 0.488   & 13.579   & 0.363  & 109.723 \\
\bottomrule
\end{tabular}
\end{table}



\section{Increasing Retrieval Chunk Size}
\label{sec:increasing} 

\begin{figure}[htbp]
    \centering    \includegraphics[width=1.00\linewidth]{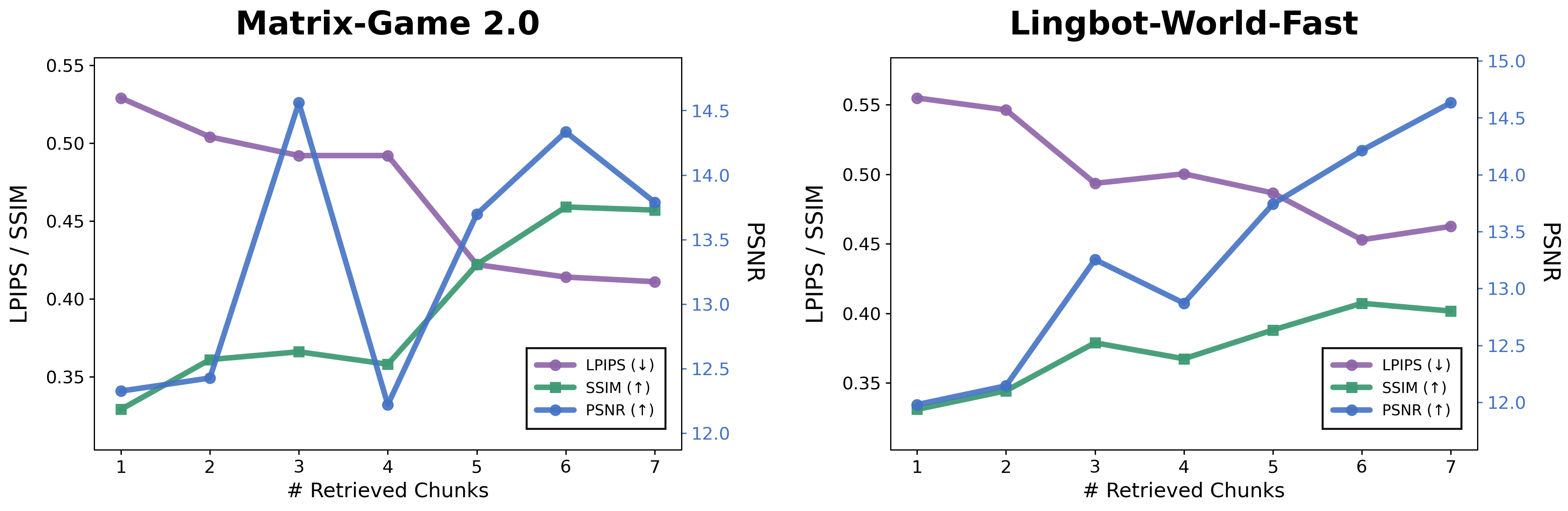}
    \caption{\textbf{Effect of increasing the number of retrieved chunks on memory fidelity.} Retrieving more past KV-cache chunks generally improves reconstruction metrics on both models, motivating World Compression as a way to fit more historical chunks within a fixed attention-window budget.
    }
    \label{fig:numchunk}
\end{figure}

Fig.~\ref{fig:numchunk} shows the effect of increasing the number of
retrieved chunks on memory fidelity. On both Matrix-Game-2.0 and
LingBot-World-Fast, LPIPS, PSNR, and SSIM generally improve as more chunks are
retrieved, indicating that broader access to historical KV caches improves
revisit consistency. This trend further motivates World Compression: beyond
reducing GPU/CPU storage cost, compression allows more historical chunks to fit
within a fixed attention-window budget, expanding retrieval coverage.

\section{WorldKV in Inspatio-World}
\label{sec:inspatio}



\begin{figure}[htbp]
    \centering
    \includegraphics[width=1.0\linewidth]{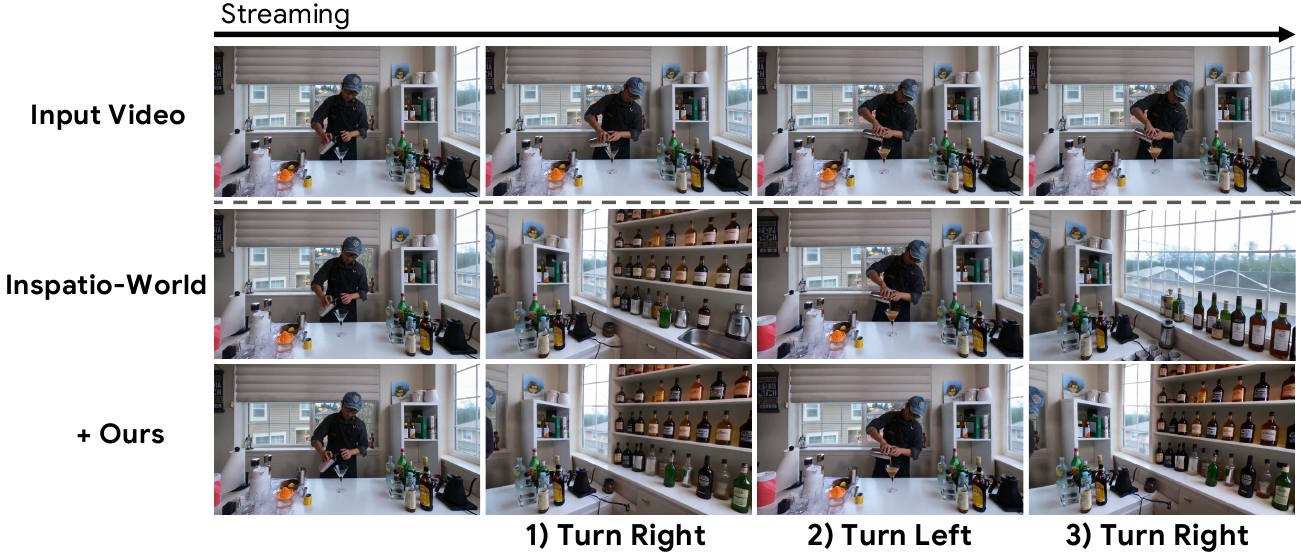}
    \caption{Applying WorldKV to Inspatio-World~\cite{inspatioteam2026inspatioworldrealtime4dworld}, a video-to-video 4D world model. Top: input video with a fixed camera. Middle: Inspatio-World generates novel-view sequences from the input video but loses scene memory upon revisit. Bottom: with WorldKV applied, the same scene content is preserved consistently across views without any fine-tuning.}
    \label{fig:inspatio}
\end{figure}

To demonstrate that WorldKV generalizes beyond the two base models used in our main evaluation, we apply it to Inspatio-World~\cite{inspatioteam2026inspatioworldrealtime4dworld}, a video-to-video 4D world model that generates novel-view sequences conditioned on an input video. Inspatio-World natively maintains memory for the input video by placing it in the sink region of the attention window, but has no mechanism to preserve memory for newly generated scenes. Like Matrix-Game-2.0~\cite{he2025matrix}, the model was not trained on long video sequences.
As shown in Fig.~\ref{fig:inspatio}, applying WorldKV enables Inspatio-World~\cite{inspatioteam2026inspatioworldrealtime4dworld} to preserve long-term memory consistency across revisits. This shows that WorldKV is not tied to specific models and can apply broadly to KV-cache-based autoregressive world models.

\section{Additional Qualitative Results}
\begin{figure}[htbp]
    \centering
    \includegraphics[width=0.97\linewidth]{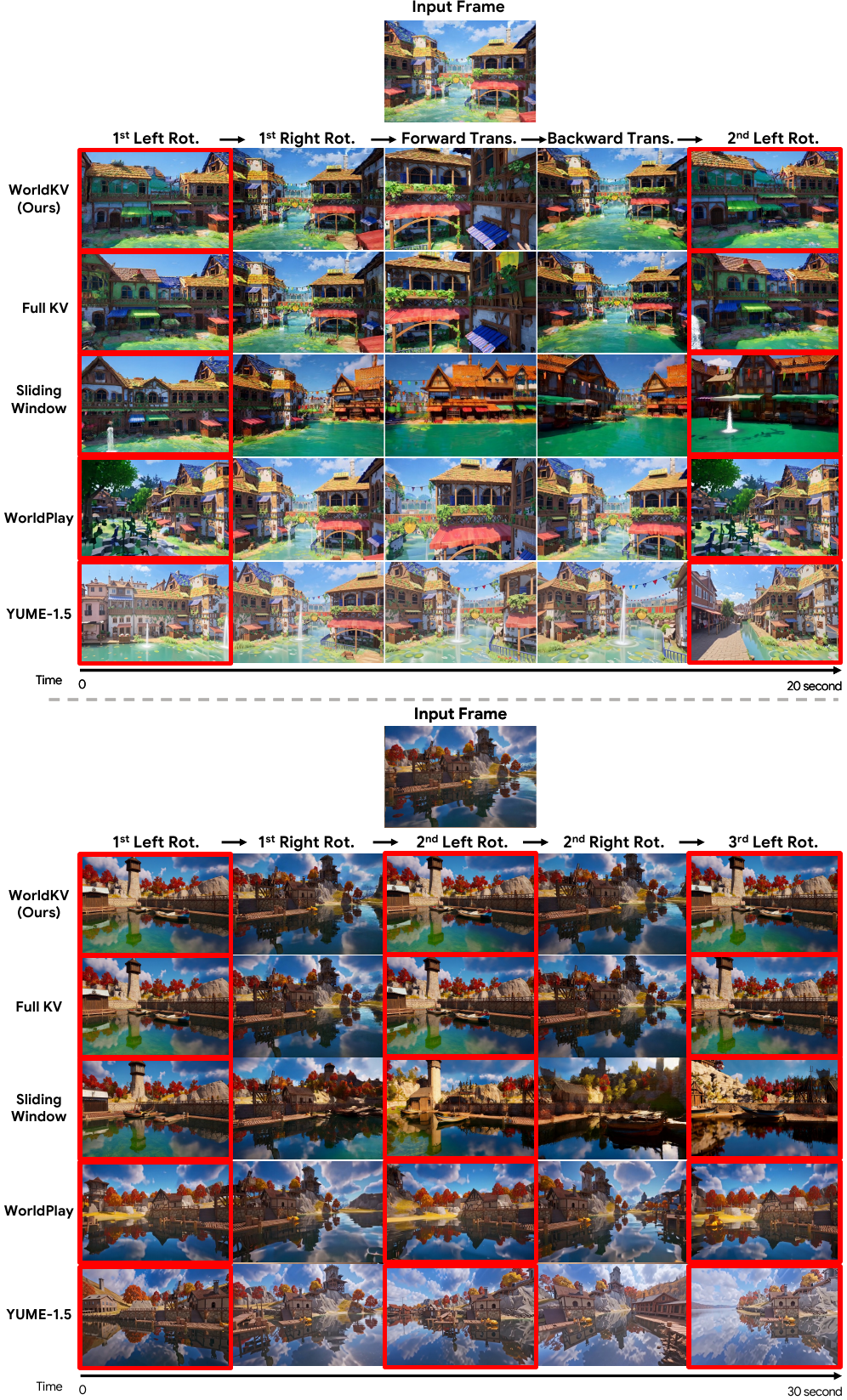}
    \caption{Frame-by-frame comparison across methods on two trajectories.}
    \label{fig:appendix_qual}
    \vspace{-2mm}
\end{figure}
Additional qualitative results of WorldKV are presented in
Fig.~\ref{fig:appendix_qual}. These examples show
that our training-free retrieval and compression framework maintains consistent
scene revisits across long-horizon trajectories, achieving qualitative results
comparable to full KV-cache attention and memory-trained baselines while
preserving efficient inference.




\section{Broader Impact}
\label{sec:impact}

This paper studies efficient long-horizon inference with long-term memory for
autoregressive video world models. By reducing the cost of maintaining and
retrieving long-term visual memory, WorldKV may improve the practicality of
interactive simulation, gaming, embodied AI, and robotic training environments,
while reducing the memory and computation required by full-history attention.
Because WorldKV is an inference-time memory-management method, it does not
directly introduce new content-generation capabilities; however, it can make
persistent interactive generation more efficient and accessible.

Potential risks therefore follow those of generative video and interactive world
models more broadly, including misuse for misleading synthetic content,
unlabeled simulated media, or more realistic persistent virtual environments.
Responsible deployment should include provenance, disclosure, and appropriate
access controls when such systems are used in public-facing applications.

\clearpage


\newpage

\end{document}